\documentclass[sigconf]{acmart}

\usepackage{graphicx}
\usepackage{textcomp}
\usepackage{adjustbox}
\usepackage{xcolor}
\usepackage{booktabs}
\usepackage{multirow}
\usepackage{makecell}
\usepackage{pifont}
\usepackage{titlesec}
\usepackage[inkscapelatex=false]{svg}

\AtBeginDocument{%
  }

\copyrightyear{2025}
\acmYear{2025}
\setcopyright{cc}
\setcctype{by}
\acmConference[WWW Companion '25]{Companion Proceedings of the ACM Web Conference 2025}{April 28-May 2, 2025}{Sydney, NSW, Australia}
\acmBooktitle{Companion Proceedings of the ACM Web Conference 2025 (WWW Companion '25), April 28-May 2, 2025, Sydney, NSW, Australia}
\acmDOI{10.1145/3701716.3717565}
\acmISBN{979-8-4007-1331-6/2025/04}

\begin{document}

\title{CubeRobot: Grounding Language in Rubik’s Cube Manipulation via Vision-Language Model}

\author{Feiyang Wang}
\authornote{Equal Contribution}

\orcid{1234-5678-9012}
\affiliation{%
  \institution{Central South University}
  \city{Changsha}
  \state{Hunan}
  \country{China}}
  \email{8207210931@csu.edu.cn}

\author{Xiaomin Yu}
\authornotemark[1]
\affiliation{%
  \institution{Great Bay University}
  \city{Guangdong}
  \country{China}}
\email{yuxm02@gmail.com}

\author{Wangyu Wu}
\authornote{Corresponding Author}
\affiliation{%
  \institution{The University of Liverpool}
  \city{Liverpool}
  \country{UK}}
\email{wangyu.wu@liverpool.ac.uk}

\renewcommand{\shortauthors}{Feiyang Wang, Xiaomin Yu, and Wangyu Wu}

\begin{abstract}
Proving Rubik's Cube theorems at the high level represents a notable milestone in human-level spatial imagination and logic thinking and reasoning. Traditional Rubik's Cube robots, relying on complex vision systems and fixed algorithms, often struggle to adapt to complex and dynamic scenarios. To overcome this limitation, we introduce CubeRobot, a novel vision-language model (VLM) tailored for solving 3x3 Rubik's Cubes, empowering embodied agents with multimodal understanding and execution capabilities. We used the CubeCoT image dataset, which contains multiple-level tasks (43 subtasks in total) that humans are unable to handle, encompassing various cube states. We incorporate a dual-loop VisionCoT architecture and Memory Stream, a paradigm for extracting task-related features from VLM-generated planning queries, thus enabling CubeRobot to independent planning, decision-making, reflection and separate management of high- and low-level Rubik’s Cube tasks. Furthermore, in low-level Rubik's Cube restoration tasks, CubeRobot achieved a high accuracy rate of 100\%, similar to 100\% in medium-level tasks, and achieved an accuracy rate of 80\% in high-level tasks.    
\end{abstract}

\ccsdesc[300]{Computer systems organization~External interfaces for robotics}

\keywords{Vision-Language Model, Multimodality, Embodied AI, Robotics and AI}


\maketitle

\section{Introduction}
\begin{figure}
  \includegraphics[width=0.5\textwidth]{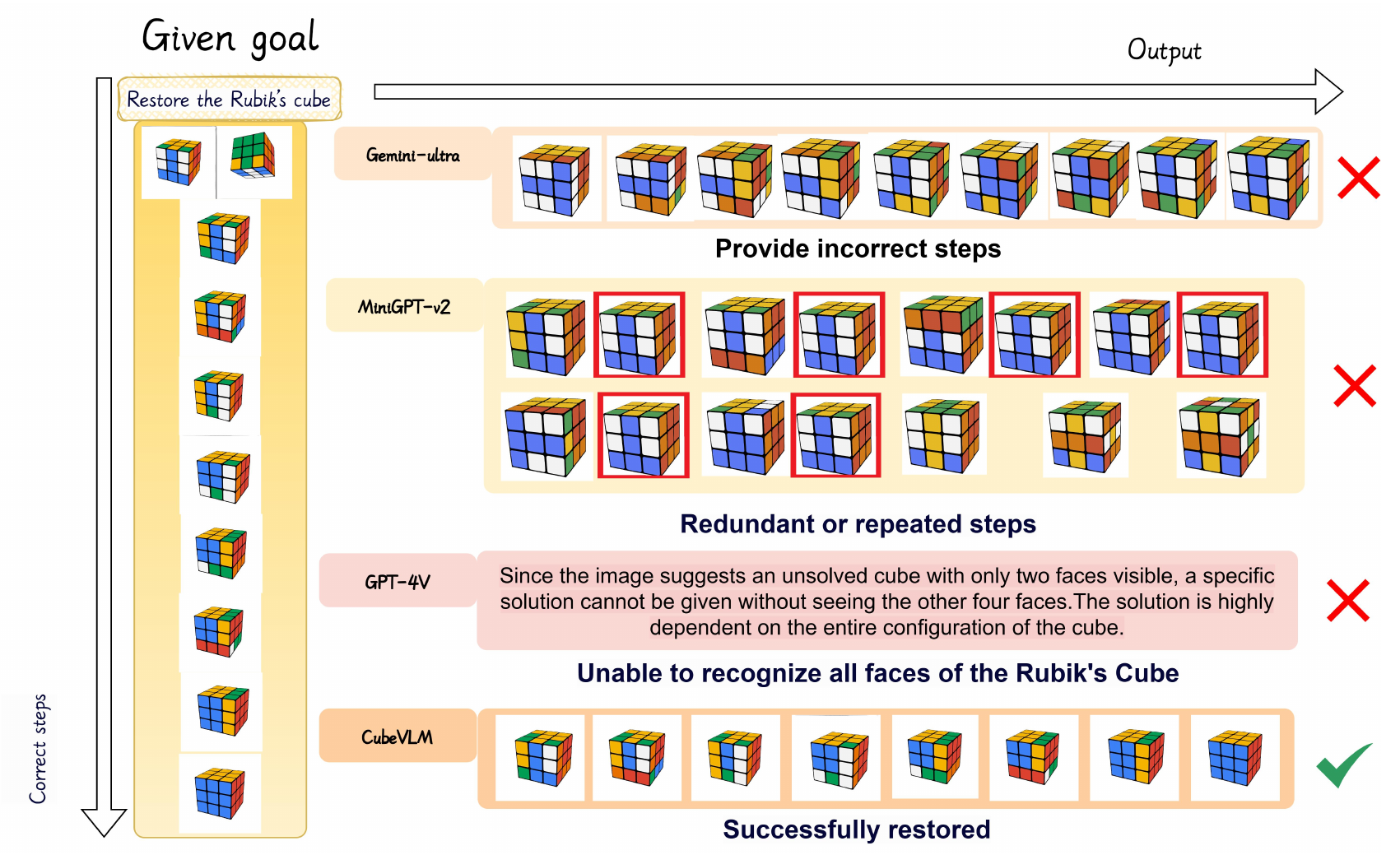}
  \caption{Comparison of Rubik's Cube Solving Performance.}
  \Description{The outputs of Gemini-ultra, MiniGPT-v2, GPT-4V, and CubeRobot in the intermediate difficulty Rubik's Cube restoration task.}
  \label{fig:t1}
  \vspace{-0.4cm}
\end{figure}

As VLMs have demonstrated remarkable performance across a wide range of natural language processing tasks \cite{mei2024gamevlm,guo2025underwater,duan2024cityllava,wu2024prompt,verma2024human,wu2025adaptive,guo2024dual-hybrid,keita2024harnessing}, we aim to explore their potential in more complex downstream applications, such as Rubik's Cube solving \cite{Yang_2020,Corli_2021,openai2019solving}. The Rubik's Cube, as a 3D puzzle, serves as a comprehensive showcase of human intelligence, patience, and spatial reasoning skills. Beyond this, our primary objective is to emulate and comprehend the human approach to problem-solving, thereby gaining insights to enhance our algorithms and models. This pursuit transcends mere technological advancement; it signifies a profound appreciation and reverence for the intricacies of human thought processes.
Existing multi-modal large language models (e.g., LLaVA \cite{liu2023improved}, Flamingo \cite{alayrac2022flamingo}, BLIP-2 \cite{li2023blip2}, PaLM-E \cite{driess2023palme}) excel in processing language and image data. However, they still face challenges in comprehending complex 3D environments, exhibiting limitations in depth perception, object relationships, and spatial reasoning(Figure. \ref{fig:t1}). Moreover, while current language models have demonstrated some long-chain reasoning capabilities, they still fall short of the capacity to independently solve highly complex tasks such as Rubik's Cube restoration.

1.We introduce CubeRobot, a Vision-Language model(VLM) tailored for solving the 3x3 Rubik's Cube.

2. We concentrate on solving the 3x3 Rubik's Cube and have compiled a specialized dataset CubeCoT for this task. Additionally, we've established high, medium, and low-difficulty tasks to evaluate the model's solving capabilities.

3. We present a dual-loop process chain architecture, consisting of an outer-loop and an inner-loop, to enhance efficiency in handling multidimensional tasks. 

4. We introduce Memory Stream, a novel memory system that logs natural language descriptions and timestamps, optimizing robots' decision-making and boosting CubeRobot's efficiency in solving Rubik's Cubes.
\section{Related Works}
\subsection{Large language models}
Recent advancements in large language models (LLMs) have been fueled by increased training data and the expansion of model parameters. Initial models like BERT~\citep{bert}, GPT-2~\citep{gpt2}, and T5~\citep{t5} laid the groundwork for this progress. The release of GPT-3~\citep{gpt3}, with its impressive 175 billion parameters, marked a pivotal moment, achieving remarkable performance across a broad spectrum of language tasks. This milestone catalyzed the development of several other large-scale models, such as Megatron-Turing NLG~\citep{smith2022using}, Chinchilla~\citep{hoffmann2022training}, PaLM~\citep{chowdhery2022palm}, OPT~\citep{zhang2022opt}, BLOOM~\citep{scao2022bloom}, and LLaMA~\citep{llama}. Moreover, Wei \textit{et al.}~\citep{wei2022emergent} identified certain \textit{emergent abilities} that manifest only in large-scale models, emphasizing the critical role of scale in the evolution of these systems. Further advancements have come with the alignment of GPT-3 through human feedback and instructions, giving rise to InstructGPT~\citep{instructGPT} and ChatGPT~\citep{chatgpt}, which enable interactive, human-like conversations and answer a wide variety of complex queries. Recently, models like Alpaca~\citep{alpaca} and Vicuna~\citep{vicuna2023}, built on LLaMA~\citep{llama}, have been open-sourced and exhibit comparable performance levels.

\subsection{LLMs in Vision-Language Tasks}
In recent years, there has been a growing trend of utilizing autoregressive language models as decoders for vision-language tasks~\cite{wu2024image,kosmos,tiong2022plug,jiang2023structure,xiao2024short,xiao2022people,xiao2023corporate,li2024real,li2024joint}, a strategy that has garnered significant attention~\citep{visualgpt,wu2023image,yang2022zero,wu2024top,jiang2022edsf,wu2025adaptive,yao2024swift,yao2023augdmc,wu2024prompt,alayrac2022flamingo,blip2,blip1,palm_e}. This method leverages cross-modal transfer, enabling shared knowledge between the language and multimodal domains. Early work, such as VisualGPT~\citep{visualgpt} and Frozen~\citep{tsimpoukelli2021multimodal}, illustrated the advantages of using a pre-trained language model as a vision-language decoder. Flamingo~\citep{alayrac2022flamingo} introduced a novel approach by combining a pre-trained vision encoder with a language model through gated cross-attention, training it on billions of image-text pairs, and demonstrating impressive few-shot learning capabilities in context. BLIP-2~\citep{blip2} later improved on this by integrating a Flan-T5~\citep{flanT5} model with a Q-Former to more efficiently align visual features with the language model. More recently, PaLM-E~\citep{palm_e}, with a staggering 562 billion parameters, advanced the integration of real-world sensory data into LLMs, bridging the gap between real-world perceptions and human language. Furthermore, the release of GPT-4~\citep{gpt4} has enhanced visual understanding and reasoning, demonstrating superior performance after training on a vast dataset of aligned image-text pairs.

Large language models (LLMs), like ChatGPT, have significantly improved the performance of vision-language tasks when combined with specialized models. Visual ChatGPT~\citep{visualChatGPT} and MM-REACT~\citep{yang2023mmreact} show how ChatGPT can act as an integrator, coordinating various visual foundation models to address more complex problems. In a similar vein, ChatCaptioner~\citep{chatcaptioner} uses ChatGPT as an interrogator, generating diverse questions that BLIP-2 answers. Through multiple conversation rounds, ChatGPT extracts visual details from BLIP-2 and summarizes the image content effectively. The Video ChatCaptioner~\citep{chen2023video} extends this idea, applying it to the spatiotemporal understanding of videos. ViperGPT~\citep{vipergpt} highlights the capability of combining an LLM with multiple vision models to solve intricate visual queries programmatically. In contrast, MiniGPT-4 directly integrates visual data with the language model, enabling it to perform various vision-language tasks without relying on external vision models.

\section{Methodology}
To tackle complex environmental perception and interaction tasks, we introduce a system integrating vision-language processing, a dual-loop Chain-of-Thought (CoT) architecture, and a memory stream mechanism. This system efficiently converts visual inputs into executable instructions and enables high-level control for specific tasks, supporting effective operation and decision-making in complex environments.

\subsection{CubeRobot: Framework}
\begin{figure}[htbp]
\centering
\includegraphics[width=0.5\textwidth]{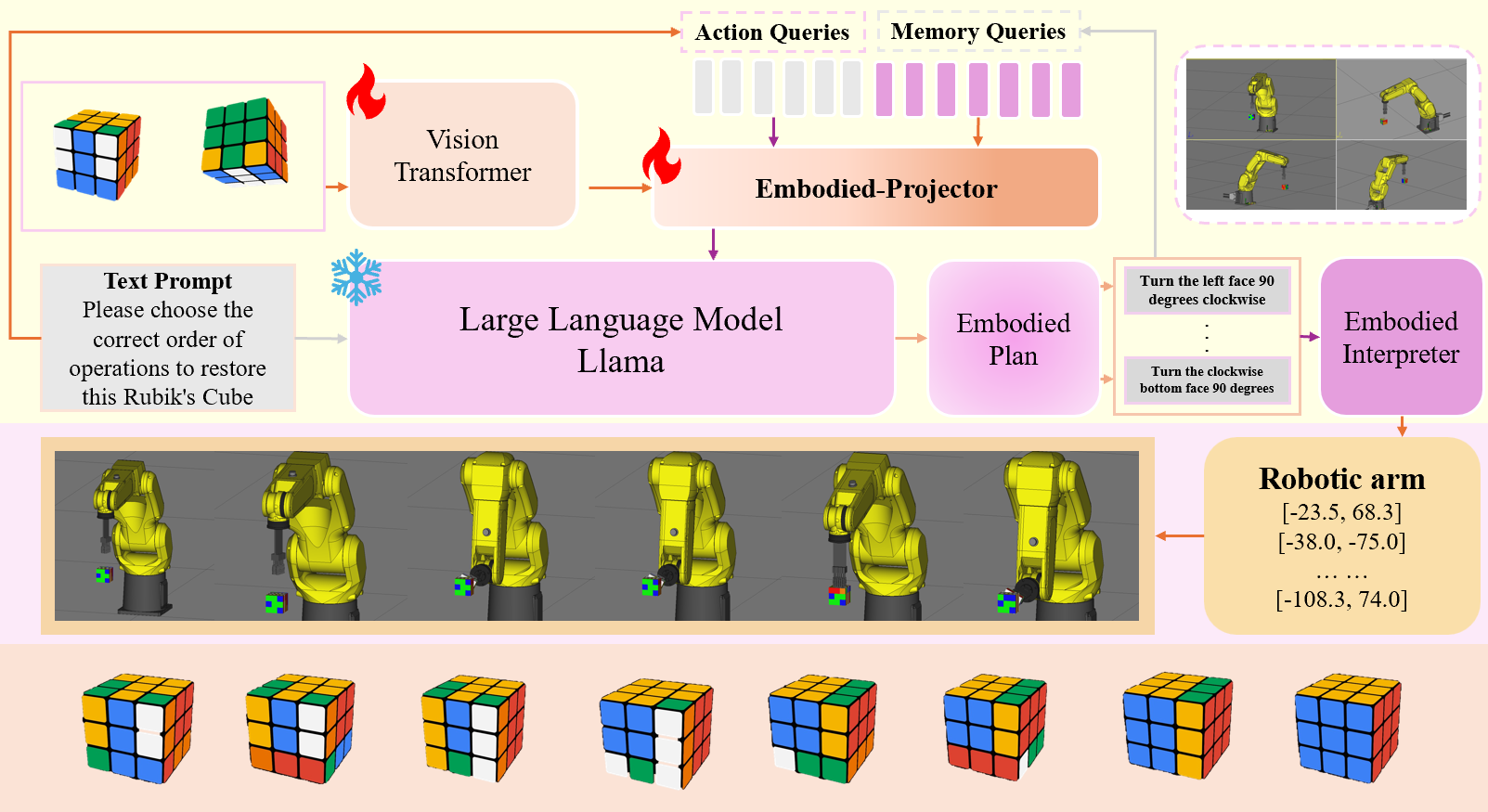}
\caption{\textbf{Framework of CubeRobot.} The orange arrow shows the vision-language planning process, while the gray arrow represents that we leverage the queried language plans for better policy learning in Rubik's Cube Manipulation tasks.}
\label{fig:5}
\vspace{-0.3cm}
\end{figure}

 Figure. \ref{fig:5} above shows the framework of CubeRobot. The Embodied-Projector, represented by the symbol $E(\cdot)$, functions as an information bottleneck that connects the frozen language model with the visual input $x_v$. It does this by supplying the language model with the most pertinent visual data. The Embodied-Projector is composed of two sub-modules: $E_v$: $x_v \to y_v$, which extracts features from the picture input, and $E_t: x_t \to y_t$, which extracts features from the text input. In order to interact with $x_v$ through cross-attention layers and with $x_t$ through self-attention layers, we use $N$ learnable Action query embeddings $y_{query}$ as the input of $E$. A mapping function represented as $M: z \to z'$, carries out this transformation using a linear projection across a fully connected (FC) layer. By acting as "soft visual prompts for the language model," the projected embeddings, $z'$, decouple the whole interaction into visual-query and query-text interactions. Using $z'$ and the text prompt as input, the language model infers the final embodied planning. The embodied plan $x_{plan}$ is utilized as the input text for Embodied-Projector to query the task-relevant instance level characteristics $z_{instance} = E(x_v, x_{plan}, y_{query})$ for high-level control, which seeks to create actions to interact with the environment. A pre-trained CubeRobot model on ViT is used to infer the global context. The output of the policy network consists of specific executable actions, such as positions and velocities in the Cartesian coordinate system.
\subsection{Dual-loop VisionCoT}

\begin{figure}[t]
\centering
\includegraphics[width=0.35\textwidth]{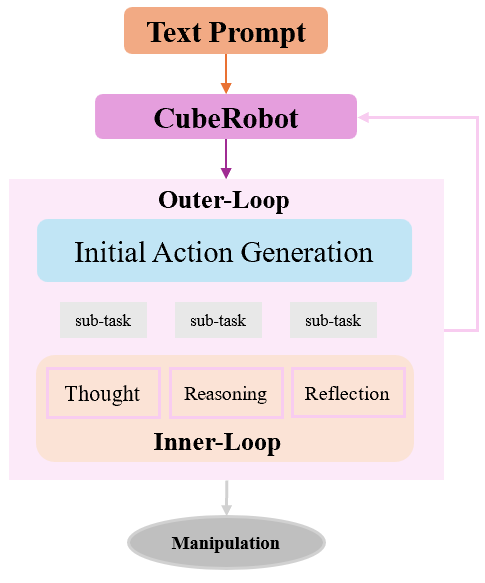}
\caption{\textbf{Dual-loop CoT.} The outer-loop manages high-level tasks, including initial action planning and iterative refinements, while tracking task progress. The inner-loop executes specific sub-tasks assigned by the outer-loop, employing thought, reasoning, and reflection.}
\label{Dual-loop CoT}
\vspace{-0.6cm}
\end{figure}

We believe that one has to consider the problem's unique components as well as its general viewpoint in order to tackle multidimensional projects efficiently \cite{xagent2023}. CubeRobot employs a dual-loop process chain architecture, as illustrated in Figure. \ref{Dual-loop CoT}, with an outer-loop process is used for high-level job management and an inner-loop process is used for low-level task execution. The outer-loop process enables the CubeRobot to recognize and segment huge assignments into smaller, more manageable sections. On the other hand, the inner-loop process acts as the painstaking executor, focusing on the particulars of the assigned responsibilities. 

Initial Action Generation: The CubeRobot starts by creating an initial plan that establishes the fundamental tactics for carrying out the assignment. This entails dividing a given difficult job $T$ into more manageable, smaller subtasks. A CubeRobot is used to break down a large task into a number of smaller tasks, $\{T_1, \cdots,T_N\}$: 

Iterative Action Refinements: CubeRobot advances by removing the first subtask from the task queue following the initial planning. The inner-loop is then given this subtask. CubeRobot keeps an eye on the state and progress of tasks all the time. The inner-loop retrieves $Reflection$ from the Memory Store and reports it following the completion of each subtask. CubeRobot initiates the proper handling procedures, such adjusting the plan or moving on to the next subtask, based on the reflection. The latest subtask's Memory Stream provides $Reflection$. Until there are no more subtasks in the queue, the outer loop is completed.

\textbf{Outer-Loop.} The outer-loop serves as the high-level planner and the primary orchestrator of tasks, acting as a supervisor for the entire problem-solving sequence. Its responsibilities can be broken down as follows:

\textbf{Inner-Loop.} The inner-loop is pivotal for executing the individual subtasks assigned by the outer-loop. Given a subtask $T_i$, an appropriate action is designated, ensuring $T_i$ reaches its intended outcome. For every action $a_t$, we integrate CubeRobot's reasoning process and the actual action into a single function call, meaning that both the action to be performed and the reasoning trail are passed as parameters of this function.

\textbf{Thought}: A brief summary of the CubeRobot's main realization of the circumstances. 

\textbf{Reasoning}: Follows the CubeRobot's logical path to the formation of its notion. 

\textbf{Reflection}: Serves as a reflection loop by recording the CubeRobot's introspection on its behaviors. It draws attention to any mistakes or possible areas for development.

\subsection{Memory Stream}

The memory stream maintains a comprehensive record of the robot’s experience. It is a list of memory objects, where each object contains a natural language description, a creation timestamp, and a most recent access timestamp. The most basic element of the memory stream is an observation, which is an event directly perceived by a robot. Common observations include behaviors performed by the robots themselves or behaviors that robots perceive as being performed by other robots or non-robot objects \cite{park2023generative}. CubeRobot assigns a higher score to memory objects that were recently accessed so that events from a moment ago or this morning are likely to remain in the robot’s attentional sphere. For instance, a mundane event, such as observing all sides of the Cube yield a high score. There are many possible implementations of an importance score; we find that directly asking the language model to output an integer score is effective. 

In our implementation, we use the language model to generate an embedding vector of the text description of each memory. Then, we calculate relevance as the cosine similarity between the memory’s embedding vector and the query memory’s embedding vector.

\begin{table*}[ht]
\centering  
\caption{Accuracy of various Visual Language Models for Rubik's Cube solving tasks}  
\label{tab:tab1}  
\setlength{\tabcolsep}{1pt}  
\begin{tabular}{cccccccccccccc}  
\toprule
\\ & llava(13b) & llava(34b) & \makecell[c]{MiniGPT4-\\Video} & \makecell[c]{blip2-flan\\-t5-xl} & \makecell[c]{blip2-flan\\-t5-xxl} & \makecell[c]{Gemini\\\-ultra} & GPT-4V & GIT & MiniGPT-V2 & MiniGPT-4 & llava(7b)& \makecell[c]{CubeRobot\\(ours)} \\
\midrule
Task finetune
&\ding{54}&\ding{54}&\ding{54}&\ding{54}&\ding{54}&\ding{54}&\ding{54}&\ding{54}&\ding{52}&\ding{52}&\ding{52}&\ding{52}\\
\toprule
low-level & 0 & 0 & 0 & 0 & 0 & 0 & 0 & 0 &73.33\% &61.11\% & 66.67\% & 100\%  \\

medium-level & 0 & 0 & 0 & 0 &0 & 0 & 0 & 0 & 22.22\% & 22.22\% & 22.22\% & 100\% \\

high-level & 0 &0 & 0 & 0 & 0 & 0 & 0 & 0 &10\% & 10\% & 10\% & 80\%  \\
\bottomrule
\end{tabular}
\vspace{-0.2cm}
\end{table*}

\begin{table} [t]
\centering  
\caption{Ablation Analysis on Rubik's Cube Tasks}  
\label{tab:tab2}  
\begin{tabular}{cccc}  
\toprule
\ Version & \ Low-level & medium-level & high-level \\
\midrule
CubeRobot & 100\% & 100\% & 80\%  \\ 

\textbf{-}Dual-loop CoT & 100\% & 38.89\% & 30\% \\ 

\textbf{-}Memory Stream& 100\% & 33.33\% & 20\% \\ 
\midrule
VLM only & 100\% & 27.78\% & 10\%\\
\bottomrule
\end{tabular}
\end{table}

\section{EXPERIMENT}

We obtained fundamental data from the website \cite{https://ruwix.com/the-rubiks-cube/rubiks-cube-patterns-algorithms/more-rubiks-patterns}, and adopted a set of 15 simple basic instructions for solving Rubik's cubes. We outlined 28 distinct Rubik's cube-solving steps and used a Rubik's cube simulator to capture images of the cube in various states.

In our study, Rubik's cube-solving tasks are divided into three difficulty levels: low, medium, and high. At the low difficulty level, there are 15 independent tasks, each solvable in a single step by executing one of our defined basic instructions. The medium difficulty level includes 18 independent tasks that require between 9 and 12 precise moves to complete cube restoration. The high difficulty level consists of 10 extremely challenging independent tasks, ranging from a minimum of 19 to a potential maximum of 31 complex moves to successfully solve the Rubik's cube.

\begin{figure}[t]
\centering
\includegraphics[width=0.5\textwidth]{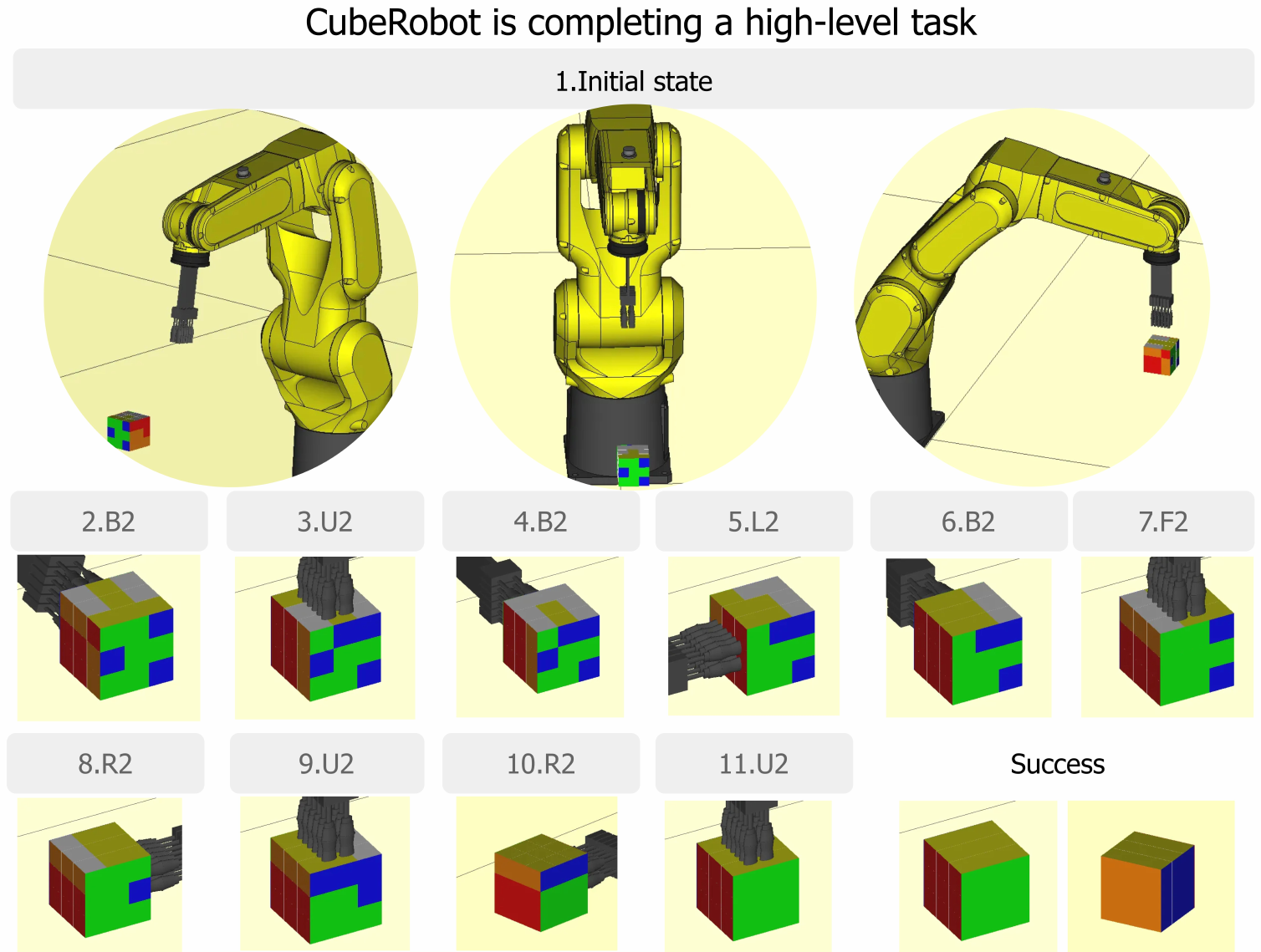}
\caption{\textbf{Visualization results of CubeRobot.} We accurately emulated the movements of the CubeRobot equipped with LR Mate 200iD robotic arm.}
\label{fig:twocolumn}
\vspace{-0.5cm}
\end{figure}

\subsection{CubeRobot Manipulation Results}
The experiments involve assessing the ability of various VLM models to solve a Rubik's cube using three tasks of increasing difficulty. By inputting the state image of the Rubik's cube along with a prompt, We obtained the accuracy of the various models. (Table.\ref{tab:tab1}).
Due to permission constraints, some Visual Language Models (VLMs) could not be fine-tuned. Notably, un-fine-tuned VLMs performed poorly in downstream Rubik's cube solving tasks, achieving 0\% accuracy across all three levels of difficulty. Specifically, in low-level tasks, models like LLaVA(34b) \cite{liu2023improved} and MiniGPT-4Video\cite{ataallah2024minigpt4videoadvancingmultimodalllms} tended to provide overly complex solutions for problems that could be solved in a single step, leading to 0\% accuracy. On the other hand, models such as Gemini\cite{geminiteam2024geminifamilyhighlycapable}, blip2 \cite{li2023blip2}, and GPT-4V \cite{yang2023dawn} failed to provide effective responses based on the given prompts. After fine-tuning, MiniGPT-V2\cite{chen2023minigptv2} showed significantly improved performance in low-strategy tasks compared to LLaVA(7b)\cite{liu2023improved} and MiniGPT-4\cite{zhu2023minigpt4}. In contrast, CubeRobot excelled in low-strategy tasks, achieving a 100\% accuracy. In more complex medium-level and high-level Rubik's Cube challenges, MiniGPT-V2 \cite{chen2023minigptv2}, LLaVA(7b) \cite{liu2023improved}, and MiniGPT-4\cite{zhu2023minigpt4} encountered deadlocks during the solution process, leading to lower accuracy. However, CubeRobot maintained its outstanding performance, achieving 100\% accuracy in the medium-level task and an impressive 80\% accuracy in the high-level task.

\subsection{Virtual Environment Simulation of CubeRobot }
We conduct simulation experiments in a virtual environment, with the assistance of the LR Mate 200iD robotic arm. Figure. \ref{fig:twocolumn} illustrates the process of CubeRobot to complete high-level tasks.

We developed an interface that translates the output of CubeRobot into a script written in the RAPID language supported by RoboGuide. This script contains specific commands and actions required for the robotic arm to solve each step of the Rubik's cube puzzle. The translation process involves converting the abstract steps generated by CubeRobot into joint angles and Cartesian coordinates that the LR Mate 200iD robotic arm can understand and execute. To monitor progress and verify the correctness of the solution, we capture real-time images after CubeRobot completes each subtask. We set up a virtual camera within RoboGuide, positioning it optimally to observe both the robotic arm and the Rubik's cube. Captured images are fed back into CubeRobot's Inner-Loop, which retrieves the "reflection" from the memory store and reports it to CubeRobot.

\subsection{Ablation study}

We conduct ablation studies to analyze the effectiveness of the inner and outer loops in the Dual-loop CoT architecture, as well as the necessity of the memory stream for model performance.

As shown in Table. \ref{tab:tab2}, the experimental results indicate that the introduction of the memory mechanism significantly improves the accuracy of the Rubik's Cube solving task compared to the Dual-loop CoT-Only model. Among the memory retrieval mechanisms, the recency, importance, and relevance scores play crucial roles in enabling the model to perform complex tasks. Moreover, the interaction between the inner and outer loops of the Dual-loop CoT architecture is essential for task decomposition and integration, which is critical for managing task execution.

\section{CONCLUSION}
\label{sec:majhead}
In this work, we present CubeRobot, an end-to-end multimodal embedded AI foundation model that enables robots to achieve the separation of high-level planning and low-level task execution. To this end, we create a Rubik's Cube dataset named CubeCoT and introduce the Dual-loop CoT , which plays a crucial role in task decomposition and integration through the interaction process between inner and outer loops.
Furthermore, this innovative approach not only streamlines task processing but also significantly improves the robot's ability to learn from and adjust to new challenges.

\bibliographystyle{ACM-Reference-Format}
\bibliography{sample-base}

\end{document}